\documentclass[sigconf, nonacm]{acmart}

\usepackage{hyperref}
\usepackage{algorithmic}
\usepackage[ruled,vlined]{algorithm2e} 
\usepackage{multirow}
\usepackage{enumitem}
\usepackage{braket}

\usepackage[most]{tcolorbox}
\tcbuselibrary{theorems,breakable}

\newtcbtheorem[auto counter]{examplebox}{Example}{
  enhanced, 
  breakable, 
  colback=blue!3, 
  colframe=blue!50!black, 
  boxrule=0.6pt, 
  arc=1mm, 
  left=2mm, 
  right=2mm, 
  top=1mm, 
  bottom=1mm, 
  fonttitle=\bfseries, 
  title style={left color=blue!15, right color=blue!5}, 
  coltitle=black
}{ex}

\begin{document}

\title[Qubit-Efficient Quantum Search for Hyperdimensional Decomposition via Logarithmic Encoding]{Qubit-Efficient Quantum Search for \\Hyperdimensional Decomposition via Logarithmic Encoding}


\author{Sanggeon Yun}
\affiliation{%
  \institution{University of California, Irvine}
  \city{Irvine}
  \state{CA}
  \country{USA}}
\email{sanggeoy@uci.edu}

\author{Hyunwoo Oh}
\affiliation{%
  \institution{University of California, Irvine}
  \city{Irvine}
  \state{CA}
  \country{USA}
}\email{hyunwooo@uci.edu}

\author{Ryozo Masukawa}
\affiliation{%
  \institution{University of California, Irvine}
  \city{Irvine}
  \state{CA}
  \country{USA}
}\email{rmasukaw@uci.edu}

\author{Raheeb Hassan}
\affiliation{%
  \institution{University of California, Irvine}
  \city{Irvine}
  \state{CA}
  \country{USA}
}\email{raheebh@uci.edu}

\author{Mohsen Imani}
\affiliation{%
  \institution{University of California, Irvine}
  \city{Irvine}
  \state{CA}
  \country{USA}
}\email{m.imani@uci.edu}


\begin{abstract}
Hyperdimensional Computing (HDC) represents symbols using high-dimensional hypervectors of dimension $D$. In hypervector decomposition, the objective is to recover $F$ constituent hypervectors, each drawn from a codebook of size $N$, from a bound target hypervector. This requires searching over $N^F$ candidate tuples, making the task computationally prohibitive at scale. Recent quantum approach provides a quadratic search advantage, but typically rely on qubit-inefficient $\mathcal{O}(D)$-qubit hypervector representations.
\
We propose a qubit-efficient quantum framework for HDC decomposition that reduces the representation cost to $\mathcal{O}(\log D)$. The framework introduces logarithmic hypervector and binding encodings, together with a reversible hypervector lookup operator for circuit-level manipulation of dense hypervectors. Combined with a modified Dürr--Høyer search procedure, the method preserves $\mathcal{O}(\sqrt{N^F})$ search complexity while substantially reducing qubit usage. Experimental results validate correct similarity computation, accurate decomposition in executable regimes, and significantly improved qubit scaling over baselines based on explicit $D$-qubit hypervector encodings, achieving up to $2{,}000\times$ fewer qubits.

\end{abstract}

\maketitle

\section{Introduction}

\begin{figure}[t]
    \centering
    \includegraphics[width=\linewidth]{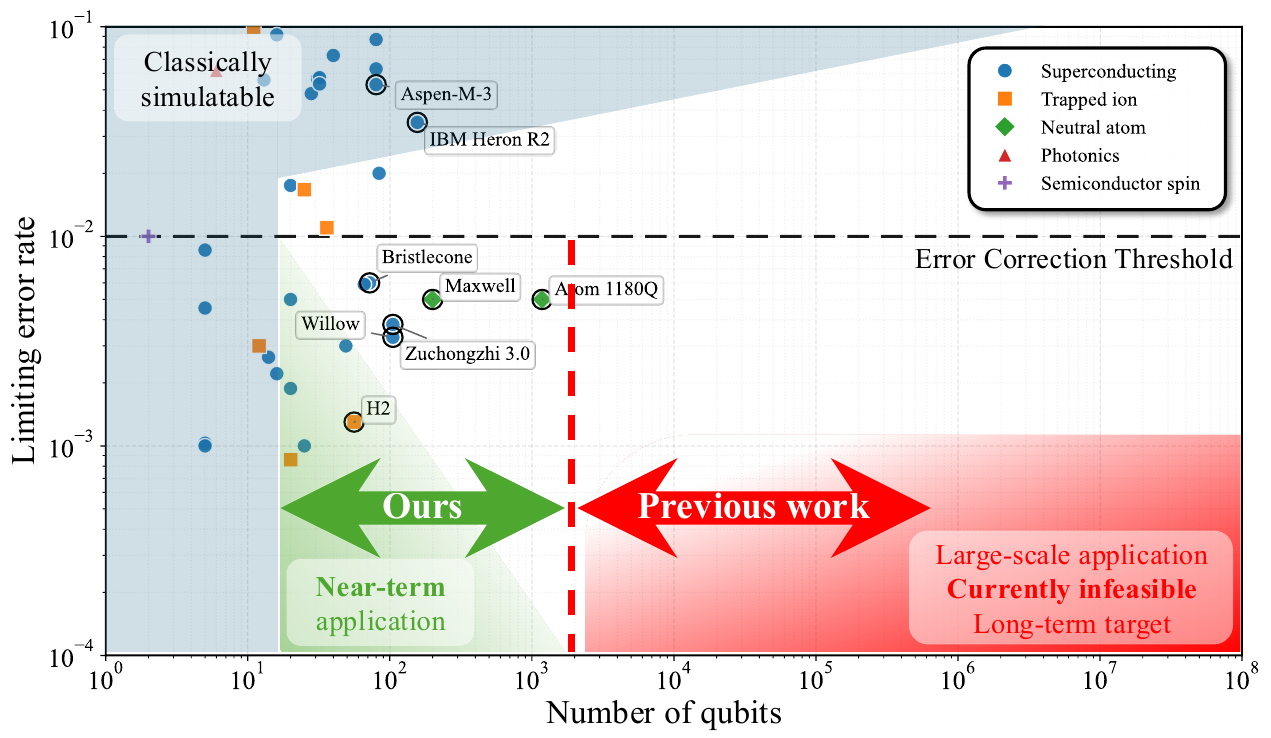}
    \vspace{-6mm}
    \caption{Current quantum hardware does not yet provide the combination of very large qubit counts and low error rates required by prior explicit-dimension HDC decomposition methods, which use \(\mathcal{O}(D)\) qubits for a \(D\)-dimensional hypervector. This motivates our logarithmic-qubit approach, which reduces the representation cost to \(\mathcal{O}(\log D)\) qubits and is better aligned with near-term hardware.}
    \label{fig:quantum_motivation}
    \vspace{-2mm}
\end{figure}

Quantum computing~\cite{nielsen2010quantum} offers a fundamentally different computational model for solving search and optimization problems by exploiting superposition, interference, and amplitude amplification~\cite{nielsen2010quantum,grover1996fast,dalzell2023quantum,boyer1998tight,brassard2000quantum}. One of its most important algorithmic advantages is the ability to reduce the complexity of unstructured search from linear to square-root scaling in the size of the search space. This makes quantum search especially attractive for machine learning problems whose dominant cost arises from combinatorial candidate selection rather than feature extraction alone~\cite{wiedemann2022quantum,xu2025accelerating,wiebe2015quantum}.

A representative instance of this opportunity arises in \emph{Hyperdimensional Computing} (HDC), also known as Vector Symbolic Architectures~\cite{kanerva2009hyperdimensional,plate1995holographic}. HDC represents information using high-dimensional vectors, called hypervectors, together with simple algebraic operations such as \emph{binding} and \emph{bundling}~\cite{kanerva2009hyperdimensional,plate1995holographic}. These operations support efficient compositional representations of structured information, including tuples, sequences, and symbolic memories, while preserving lightweight computation in high-dimensional spaces. HDC has attracted renewed interest not only as a standalone computing paradigm, but also as a structured layer on top of deep learning systems. In such hybrid models, deep neural networks (DNNs) serve as powerful feature extractors, whereas HDC provides more interpretable and explicitly compositional structure over the resulting representations~\cite{rahimi2017high,neubert2021hyperdimensional,yun2024neurohash,barkam2023reliable,lee2023comprehensive}. This integration is particularly attractive in settings where the representational power of DNNs is desired together with the symbolic structure and controllability offered by HDC.

A key obstacle, however, lies in \emph{decomposition}: recovering the constituent hypervectors from a composite representation. This problem is especially important in hybrid DNN--HDC systems, where one may wish to interpret, manipulate, or retrieve the symbolic factors encoded in a learned hypervector~\cite{memisevic2010learning,frady2020resonator,yun2025decohd,pilligua2025hypernvd}. If a target hypervector is formed by binding \(F\) factors and each factor is drawn from a codebook of size \(N\), then exact recovery requires searching over \(N^F\) possible tuples. The resulting exponential search space rapidly becomes the dominant computational bottleneck, even though the forward HDC operations themselves remain simple~\cite{kleyko2023survey,poduval2024hdqmf}.

A recent effort to address this challenge is Hyperdimensional Quantum Maximum Finding (HDQMF)~\cite{poduval2024hdqmf}, which applies a Grover-style quantum search strategy to HDC decomposition and achieves the expected quadratic improvement from \(\mathcal{O}(N^F)\) to \(\mathcal{O}(\sqrt{N^F})\)~\cite{grover1996fast}. While promising, two important limitations remain. First, \textbf{the prior formulation is \emph{qubit-inefficient}}: it represents a \(D\)-dimensional hypervector using \(\mathcal{O}(D)\) qubits, which is incompatible with the high-dimensional regime in which HDC is typically deployed, often with \(D \gtrsim 10^4\). At the same time, scaling the number of reliably controllable entangled qubits remains a major challenge in quantum hardware~\cite{kelly2018bristlecone,google2023suppressing,bausch2024learning}, and current state-of-the-art reliable processors still operate only at roughly the \(10^2\)-qubit scale~\cite{google2025quantum}. Consequently, qubit-efficient hyperdimensional methods are essential for realizing practically meaningful quantum advantage, as illustrated in \autoref{fig:quantum_motivation}. Second, \textbf{the prior formulation does not provide a detailed gate-level realization of how dense hypervectors are loaded and manipulated within the circuit}, instead treating hypervector access at a conceptual level. These limitations make it difficult to assess whether the approach can be implemented concretely while retaining meaningful scalability.

In this work, we propose a \textbf{qubit-efficient and circuit-realizable quantum framework for HDC decomposition} that addresses both limitations while preserving the same asymptotic search complexity as prior quantum work. To resolve the qubit-efficiency issue, we introduce \emph{logarithmic hypervector encoding} and \emph{logarithmic binding encoding}, which represent and manipulate \(D\)-dimensional hypervectors using only \(\mathcal{O}(\log D)\) qubits rather than \(\mathcal{O}(D)\). To resolve the circuit-realization issue, we introduce a \emph{hypervector lookup operator} that provides an explicit gate-level mechanism for retrieving and applying dense hypervector coordinates within the circuit. Building on these components, we formulate a modified Dürr--Høyer quantum search procedure~\cite{durr1996quantum} for fast HDC decomposition, retaining the \(\mathcal{O}(\sqrt{N^F})\) search complexity of prior quantum approaches while using substantially fewer qubits.

We evaluate the proposed framework through circuit-level validation and decomposition experiments across a range of problem settings. The results show that the proposed circuits compute the intended HDC similarity score correctly, recover the correct decomposition in executable regimes, and achieve substantially improved qubit scaling relative to explicit hypervector encoding-based quantum baselines.

Our main contributions are summarized as follows:
\begin{enumerate}
\item We introduce \textbf{logarithmic hypervector encoding and logarithmic binding encoding} for quantum HDC decomposition, \textbf{reducing hypervector representation cost from \(\mathcal{O}(D)\) qubits to \(\mathcal{O}(\log D)\) qubits}.
\item We propose a \textbf{hypervector lookup operator} that provides an explicit circuit-level realization of dense hypervector access and manipulation.
\item We develop a \textbf{modified Dürr--Høyer quantum search framework for HDC decomposition} that preserves the quadratic search improvement of prior quantum approaches while operating with substantially fewer qubits.
\item We \textbf{empirically validate the proposed method} through circuit-level correctness experiments, decomposition accuracy studies, and qubit-scaling analysis.
\end{enumerate}

\section{Preliminaries}

\subsection{Hyperdimensional Computing}

\noindent\textbf{HDC basics.}
HDC is a representation and computation paradigm that encodes information using high-dimensional distributed vectors, called \emph{hypervectors}~\cite{kanerva2009hyperdimensional,plate1995holographic}, typically with dimensionality $D$ in the thousands or tens of thousands. A hypervector may represent a symbol, an attribute, or a structured object. A central property of HDC is that complex structures can be constructed through a small set of algebraic operations while preserving computational simplicity. In this work, we focus on the standard bipolar setting, where each hypervector is given by $h\in\{-1, +1\}^D$.
\
Two fundamental HDC operations are \emph{binding} and \emph{bundling}. In this work, we denote binding by \(\odot\). Binding combines multiple hypervectors \(h_1,h_2,\dots\) into a single composite hypervector \(h\) according to
\begin{equation}
h = h_1 \odot h_2 \odot \cdots,
\end{equation}
which, in the bipolar setting considered here, is implemented by elementwise multiplication,
\begin{equation}
h[u] = \prod_j h_j[u],
\end{equation}
where \(h[u]\) denotes the \(u\)-th coordinate of \(h\), with \(u\in\{0,1,\dots,D-1\}\). Bundling aggregates multiple hypervectors into a single memory-like representation~\cite{kanerva2009hyperdimensional,plate1995holographic}, typically through elementwise addition followed by an optional normalization or thresholding step. To compare such composed hypervectors, we measure similarity by the normalized bipolar correlation
\begin{equation}
\delta(a,b)=\frac{1}{D}\sum_{u=0}^{D-1} a[u]\,b[u],
\end{equation}
which takes values in \([-1,1]\), equals \(1\) when \(a=b\), and is typically close to \(0\) for unrelated hypervectors. Together, these operations allow HDC systems to represent structured objects such as tuples, sequences, and graphs while retaining low computational cost.

\medskip\noindent\textbf{HDC decomposition problem.}
Although HDC operations make information composition highly efficient, recovering the underlying factors from a composite representation is computationally difficult. In particular, if a composite hypervector is constructed by binding \(F\) factors and each factor is selected from a codebook of size \(N\), then exact decomposition requires searching over \(N^F\) possible tuples in the worst case. Thus the search space grows exponentially with the number of factors, yielding a computational complexity of \(\mathcal{O}(N^F)\) before accounting for the cost of evaluating the similarity of each candidate. In this work, we focus on the decomposition problem for bound hypervectors.

\subsection{Quantum Computing}

\noindent\textbf{Quantum computing basics.}
Quantum computing~\cite{nielsen2010quantum} provides a computational model based on the manipulation of quantum states through unitary operations. The fundamental unit of quantum information is the \emph{qubit}, whose state lies in a two-dimensional complex vector space spanned by the computational basis states \(\ket{0}\) and \(\ket{1}\). An \(n\)-qubit system lies in a \(2^n\)-dimensional Hilbert space~\cite{vourdas2004quantum}, enabling a compact representation of exponentially large state spaces using only a linear number of physical qubits. Computation is performed by applying quantum gates, which are unitary transformations acting on one or more qubits. By exploiting superposition and interference, quantum circuits can amplify amplitudes corresponding to desirable solutions while suppressing others. A canonical example is Grover's search algorithm~\cite{grover1996fast}, which achieves a quadratic speedup for unstructured search.

\medskip\noindent\textbf{Qubit scalability.}
A central challenge in practical quantum computing is the limited scalability of reliable qubit systems~\cite{kelly2018bristlecone,google2023suppressing,bausch2024learning}. Unlike classical bits, qubits must preserve coherent quantum states and, when entangled, jointly occupy a tensor-product Hilbert space. The difficulty of maintaining coherence, controlling multi-qubit interactions, and suppressing noise increases rapidly with system size. As a result, quantum algorithms that require large numbers of qubits are often impractical on current hardware~\cite{kelly2018bristlecone,google2025quantum}. This limitation is especially severe for algorithms that require a large number of qubits.

\medskip\noindent\textbf{Quantum HDC decomposition.}
Recent work introduced Hyperdimensional Quantum Maximum Finding~\cite{poduval2024hdqmf}, which applies a modified Grover-style quantum search procedure to the HDC decomposition problem. While that approach reduces the search complexity from \(\mathcal{O}(N^F)\) to \(\mathcal{O}(\sqrt{N^F})\), it does so using what we refer to as \textbf{explicit hypervector encodings}, in which each hypervector dimension is represented by a separate qubit. For realistic HDC regimes, where \(D\) is often on the order of \(10^4\), this qubit requirement is prohibitively large relative to currently available quantum hardware~\cite{google2025quantum}, whose commercially accessible systems are still typically in the low-hundreds-of-qubits regime rather than the tens-of-thousands required by such encodings. In this work, we instead develop a qubit-efficient quantum HDC decomposition method that uses \(\mathcal{O}(\log D)\) qubits for hypervector representation rather than \(\mathcal{O}(D)\), while preserving the ability to perform decomposition through quantum search.

\section{Problem Formulation}

We consider the problem of \emph{codebook-constrained hypervector decomposition}. Let
\begin{equation}
C_f=\{h_{f,0},h_{f,1},\dots,h_{f,N-1}\}\subset\{-1,+1\}^{D},
\quad f\in\{1,\dots,F\},
\end{equation}
denote the \(f\)-th codebook, where \(F\) is the number of factors, \(N\) is the number of candidate hypervectors in each codebook, and \(D\) is the hypervector dimension.
A candidate decomposition is specified by a tuple
\begin{equation}
\mathbf n=(n_1,\dots,n_F)\in\{0,1,\dots,N-1\}^F.
\end{equation}
Given such a tuple, the corresponding composite hypervector is defined by the HDC binding operation
\begin{equation}
b_{\mathbf n}
=
h_{1,n_1}\odot h_{2,n_2}\odot \cdots \odot h_{F,n_F},
\end{equation}
where \(\odot\) denotes elementwise bipolar multiplication. Equivalently, for each coordinate \(u\in\{0,1,\dots,D-1\}\),
\begin{equation}
b_{\mathbf n}[u]
=
\prod_{f=1}^{F} h_{f,n_f}[u].
\end{equation}

We assume that a target hypervector \(y\in\{-1,+1\}^{D}\) is generated by binding one entry from each codebook, possibly followed by mild corruption or noise. The decomposition task is to recover the generating tuple by solving
\begin{equation}
\mathbf n^\star
=
{\arg\max}_{\mathbf n\in\{0,1,\dots,N-1\}^F}\delta(y,b_{\mathbf n}),
\end{equation}
where
\begin{equation}
\delta(y,b_{\mathbf n})
=
\frac{1}{D}\sum_{u=0}^{D-1} y[u]\,b_{\mathbf n}[u]
\end{equation}
is the normalized bipolar similarity between the target hypervector and the candidate composite hypervector.
This formulation makes the computational bottleneck explicit. A direct classical method must evaluate $\delta(y,b_{\mathbf n})$ for every candidate tuple $\mathbf n \in \{0,1,\dots,N-1\}^F$. With $N^F$ possible tuples and a fixed hypervector dimensionality $D$, the total runtime is
\begin{equation}
\mathcal{O}(N^F D) = \mathcal{O}(N^F),
\end{equation}
where the second equality follows because $D$ is treated as a constant.
Thus, the primary computational challenge in hypervector decomposition is the exponential growth in the number of candidate tuples with respect to the number of factors $F$.

\section{Methodology}

\begin{figure}[t]
    \centering
    \includegraphics[width=\linewidth]{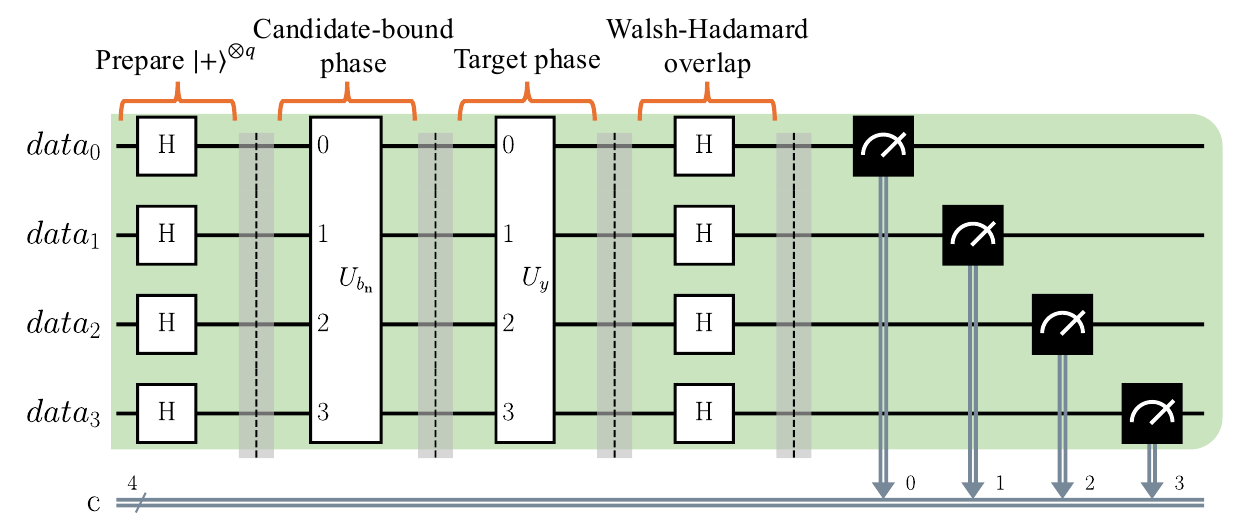}
    \vspace{-6mm}
    \caption{Similarity-extraction circuit used to compute the HDC similarity score. The data register is initialized in \(\ket{+}^{\otimes q}\), the candidate-bound phase oracle \(U_{b_{\mathbf{n}}}\) and target phase oracle \(U_y\) are applied, and a final Walsh--Hadamard transform maps the normalized bipolar similarity \(\delta(y,b_{\mathbf{n}})\) into the amplitude of the basis state \(\ket{0^q}\). Measurement of the data register therefore reveals the overlap structure needed for score evaluation.}
    \label{fig:overlap_extraction}
    \vspace{-3mm}
\end{figure}

\begin{figure*}[t]
    \centering
    \includegraphics[width=0.85\linewidth]{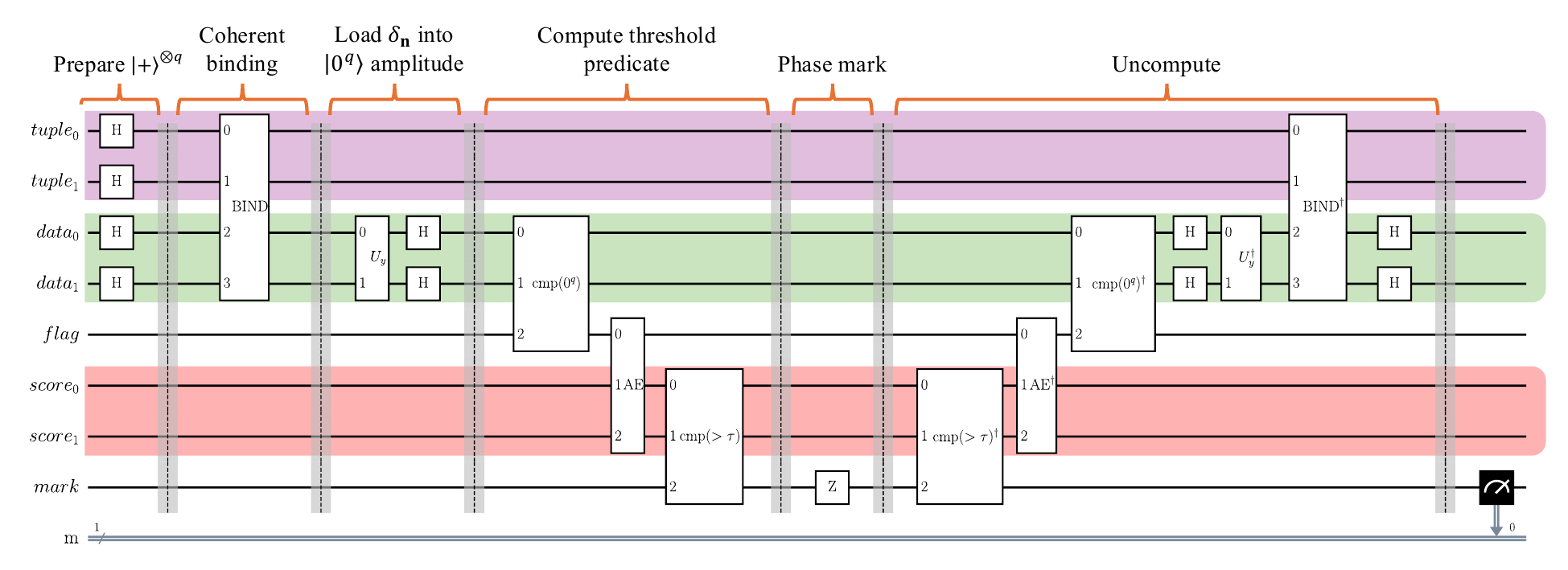}
    \vspace{-4mm}
    \caption{Compute--phase--uncompute threshold oracle used inside the modified Dürr--Høyer maximum-finding procedure. Starting from a superposition over tuple candidates and hypervector coordinates, the circuit performs coherent binding, applies the target-dependent similarity extraction, computes whether the candidate score exceeds the current threshold \(\tau\), applies a phase flip to marked tuples, and then uncomputes all work registers. This reversible structure is required so that amplitude amplification acts only on the tuple register without leaving residual garbage entanglement.}
    \label{fig:threshold_oracle}
    \vspace{-3mm}
\end{figure*}

\begin{figure}[t]
    \centering
    \includegraphics[width=\linewidth]{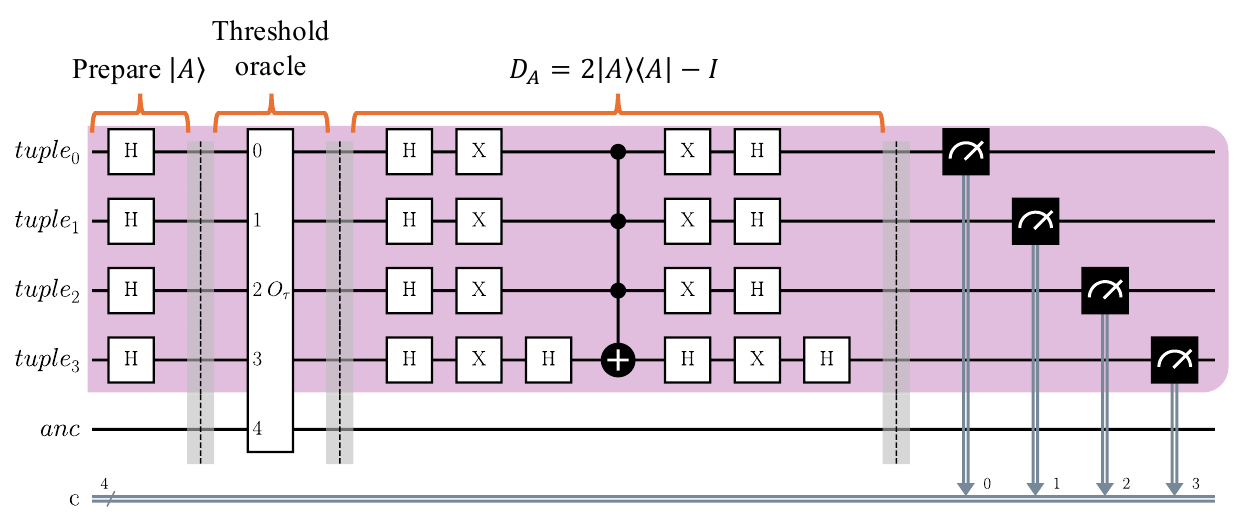}
    \vspace{-6mm}
    \caption{One amplitude-amplification step used inside the modified Dürr--Høyer maximum-finding routine. The threshold oracle \(O_\tau\) marks tuples whose score exceeds the current incumbent threshold, and the diffusion operator \(D_A = 2\ket{A}\!\bra{A} - I\) amplifies their amplitudes relative to non-improving tuples. This circuit should be interpreted as an internal subroutine of maximum finding rather than as a standalone search algorithm.}
    \label{fig:grover_iteration}
    \vspace{-3mm}
\end{figure}

\begin{figure}[t]
    \centering
    \includegraphics[width=\linewidth]{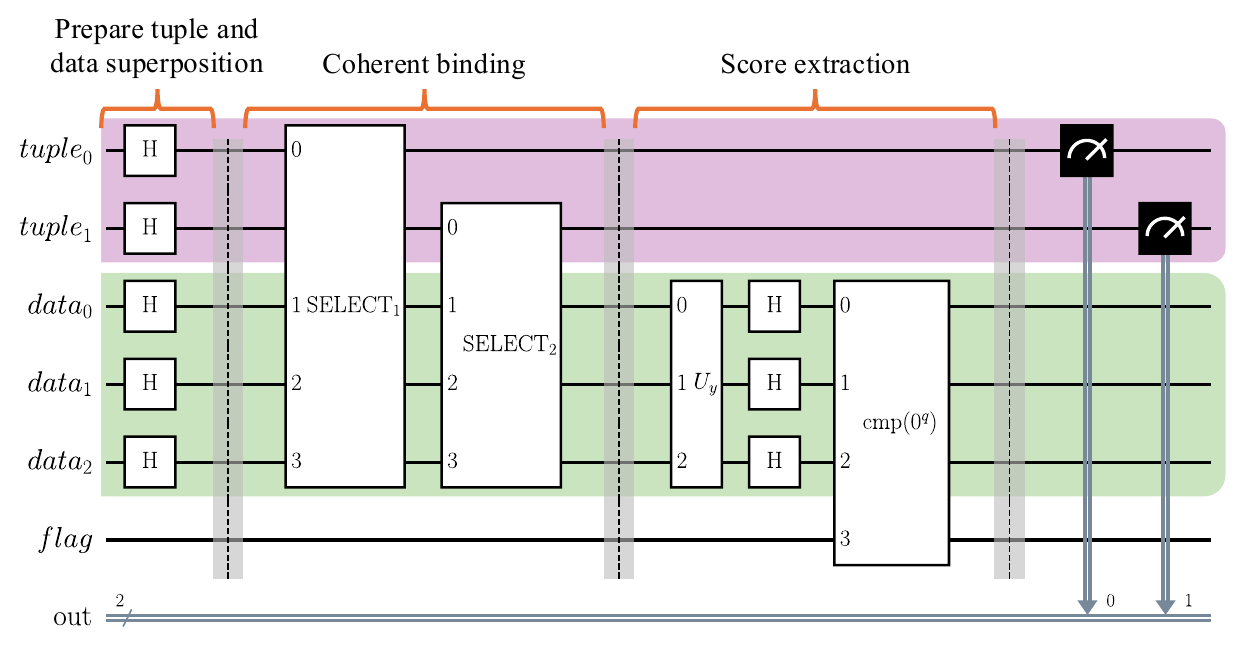}
    \vspace{-6mm}
    \caption{Tiny two-factor example illustrating the overall structure of the proposed framework. The tuple register is prepared in superposition, controlled selection operators \(\mathrm{SELECT}_1\) and \(\mathrm{SELECT}_2\) synthesize the candidate bound hypervector on the shared data register, and the target-dependent similarity computation extracts the score by concentrating amplitude on \(\ket{0^q}\). The figure provides a minimal concrete instance of coherent binding followed by score extraction.}
    \label{fig:tiny_example}
    \vspace{-3mm}
\end{figure}

We now describe the proposed quantum framework. The central idea is to separate the combinatorial search space of candidate tuples from the coordinate space of the hypervectors themselves. Unlike prior approaches based on \textbf{explicit hypervector encodings}, which allocate one qubit per hypervector dimension, we introduce what we call \textbf{logarithmic hypervector encodings} with \textbf{logarithmic binding encodings}. In this representation, the tuple search is carried by one register, while the $D$-dimensional hypervectors are represented on a shared coordinate register using only $\mathcal{O}(\log D)$ qubits. Similarity is extracted through a similarity computation that maps the HDC score into the amplitude of a distinguished basis state. This score is then used inside a modified Dürr--Høyer maximum-finding routine.

\subsection{Overall Architecture}

The quantum state is organized into three logical parts. The tuple register stores the candidate tuple $\mathbf n$ using the proposed logarithmic binding encoding, while the data register stores the associated $D$-dimensional hypervector using the proposed logarithmic hypervector encoding. Ancillary qubits support reversible lookup, phase kickback, score comparison, and threshold marking, which will be introduced later subsection in detail. If $D=2^q$, then the data register contains $q=\log_2 D$ qubits. The tuple register contains $F\lceil \log_2 N\rceil$ qubits. 
Operationally, the method proceeds as follows. The tuple register is prepared in superposition over all candidate tuples. Conditioned on each tuple branch, the circuit synthesizes the corresponding bound hypervector on the shared data register. A target-dependent similarity computation then maps the HDC similarity score into the amplitude of $\ket{0^q}$. This score is converted into a threshold predicate and used within modified Dürr--Høyer maximum finding. \autoref{fig:tiny_example} illustrates the tuple-conditioned binding stage, \autoref{fig:overlap_extraction} shows the similarity extraction stage, and \autoref{fig:threshold_oracle} and \autoref{fig:grover_iteration} show the search subroutine.

\subsection{Logarithmic Hypervector and Binding Encodings}

\noindent\textbf{Logarithmic hypervector encoding.}
Let \(D=2^q\). Instead of storing a hypervector explicitly across \(D\) qubits, we encode it as a phase pattern over the computational basis of a \(q\)-qubit coordinate register:
\begin{equation}
\ket{h}
=
\frac{1}{\sqrt D}\sum_{u=0}^{D-1} h[u]\ket{u}.
\end{equation}
All basis states have equal magnitude \(1/\sqrt D\), while the sign \(h[u]\in\{-1,+1\}\) is encoded in the phase. Equivalently, define the diagonal phase oracle
\begin{equation}
U_h\ket{u}=h[u]\ket{u},
\quad
U_h=\mathrm{diag}(h[0],h[1],\dots,h[D-1]).
\end{equation}
Applying \(U_h\) to \(\ket{+}^{\otimes q}\) produces the encoded state:
\begin{equation}
U_h\ket{+}^{\otimes q}
=
\frac{1}{\sqrt D}\sum_{u=0}^{D-1} h[u]\ket{u}.
\end{equation}
This reduces the qubit cost of representing a \(D\)-dimensional hypervector from \(\mathcal{O}(D)\) to \(\mathcal{O}(\log D)\).

\begin{examplebox}{Logarithmic-qubit hypervector encoding}{log-encoding}
Suppose \(D=8\), so the data register has \(q=3\) qubits. Let
\[
h=[+1,-1,+1,-1,-1,+1,+1,-1].
\]
Then
\[
\ket{h}
=
\frac{1}{\sqrt 8}
\bigl(
\ket{0}
-\ket{1}
+\ket{2}
-\ket{3}
-\ket{4}
+\ket{5}
+\ket{6}
-\ket{7}
\bigr).
\]
Thus a dense eight-dimensional hypervector is represented using only three qubits.
\end{examplebox}

\medskip\noindent\textbf{Logarithmic binding encoding.}
Binding in bipolar hypervectors is multiplicative in each hyperdimension. If
\begin{equation}
b=a^{(1)}\odot a^{(2)}\odot\cdots\odot a^{(F)},
\end{equation}
then for every dimension \(u\),
\begin{equation}
b[u]
=
\prod_{f=1}^{F} a^{(f)}[u].
\end{equation}
Under the proposed logarithmic hypervector encoding, this implies that the phase oracle of the bound hypervector is simply the product of the phase oracles of its factors:
\begin{equation}
U_b
=
\prod_{f=1}^{F} U_{a^{(f)}}.
\end{equation}
Since all factor oracles are diagonal in the same computational basis, the multiplication is exact and introduces no additional representational overhead. Therefore, for any fixed candidate tuple \(\mathbf n=(n_1,\dots,n_F)\), the corresponding bound hypervector can be represented as
\begin{equation}
\ket{b_{\mathbf n}}
=
U_{b_{\mathbf n}}\ket{+}^{\otimes q},
\qquad
U_{b_{\mathbf n}}
=
\prod_{f=1}^{F} U_{f,n_f}.
\end{equation}
This defines the logarithmic binding encoding for an individual candidate tuple.

\subsection{Hypervector Lookup Operator}

The logarithmic hypervector encoding described above assumes coherent access to individual hypervector coordinates $h_{f,j}[u]$ for each codebook entry $h_{f,j}$. To realize this access within a circuit model, we introduce a \textbf{hypervector lookup operator}, which retrieves the sign of the $u$-th coordinate of the selected codebook hypervector and converts it into a quantum-computable form. 
For each factor $f$, define the binary lookup table
\begin{equation}
T_f(j,u)
=
\begin{cases}
0, & h_{f,j}[u]=+1,\\
1, & h_{f,j}[u]=-1.
\end{cases}
\end{equation}
This is a classical sign table for the codebook. The corresponding reversible lookup operator is
\begin{equation}
L_f:\ket{j}\ket{u}\ket{b}
\mapsto
\ket{j}\ket{u}\ket{b\oplus T_f(j,u)}.
\end{equation}
At the matrix level,
\begin{equation}
L_f
=
\sum_{j=0}^{N-1}\sum_{u=0}^{D-1}
\ket{j,u}\!\bra{j,u}\otimes X^{T_f(j,u)}.
\end{equation}

When the ancilla is initialized in \(\ket{-}=(\ket{0}-\ket{1})/\sqrt2\), phase kickback converts the Boolean lookup into the desired sign phase:
\begin{equation}
L_f\ket{j}\ket{u}\ket{-}
=
(-1)^{T_f(j,u)}\ket{j}\ket{u}\ket{-}
=
h_{f,j}[u]\ket{j}\ket{u}\ket{-}.
\end{equation}
Thus the lookup operator induces the exact controlled phase oracle
\begin{equation}
\widetilde U_f
=
\sum_{j=0}^{N-1}\ket{j}\!\bra{j}\otimes U_{f,j},
\end{equation}
where
\begin{equation}
U_{f,j}
=
\mathrm{diag}\bigl(h_{f,j}[0],h_{f,j}[1],\dots,h_{f,j}[D-1]\bigr).
\end{equation}

\begin{examplebox}{Dense lookup for a tiny codebook}{denselookup}
Let \(N=2\) and \(D=4\), with
\[
h_{0}=[+1,-1,+1,-1],
\qquad
h_{1}=[-1,-1,+1,+1].
\]
Then
\[
T=
\begin{bmatrix}
0 & 1 & 0 & 1\\
1 & 1 & 0 & 0
\end{bmatrix},
\]
and the lookup induces
\[
\widetilde U
=
\ket{0}\!\bra{0}\otimes \mathrm{diag}(+1,-1,+1,-1)
+
\ket{1}\!\bra{1}\otimes \mathrm{diag}(-1,-1,+1,+1).
\]
This example shows how arbitrary dense codebooks are supported.
\end{examplebox}

\subsection{Coherent Binding Over Candidates}

Having defined the logarithmic binding encoding for a fixed tuple, we now show how to construct it coherently across the entire candidate space. The key idea is to place the tuple register in superposition and apply tuple-controlled phase operators on a shared data register. This yields a joint quantum state in which each tuple branch is paired with the logarithmic hypervector encoding of its corresponding bound hypervector.
\
The tuple register is initialized in the uniform superposition
\begin{equation}
\ket{A}
=
\frac{1}{\sqrt{N^F}}
\sum_{\mathbf n\in\{0, 1, \dots, N-1\}^F}
\ket{\mathbf n},
\qquad
\ket{\mathbf n}=\ket{n_1}\cdots\ket{n_F},
\end{equation}
and the data register is initialized in \(\ket{+}^{\otimes q}\). For each factor \(f\), define the controlled selection operator
\begin{equation}
\mathrm{SELECT}_f
=
\sum_{j=0}^{N-1}
\ket{j}\!\bra{j}_{n_f}\otimes U_{f,j}.
\end{equation}
The full tuple-conditioned binding operator is
\begin{equation}
\mathrm{BIND}
=
\prod_{f=1}^{F}\mathrm{SELECT}_f.
\end{equation}
Because all \(U_{f,j}\) are diagonal on the shared data register, the \(\mathrm{SELECT}_f\) commute. Applying \(\mathrm{BIND}\) yields
\begin{equation}
\mathrm{BIND}\bigl(\ket{A}\otimes \ket{+}^{\otimes q}\bigr)
=
\frac{1}{\sqrt{N^F}}
\sum_{\mathbf n\in\{0, 1, \dots, N-1\}^F}
\ket{\mathbf n}\ket{b_{\mathbf n}}.
\end{equation}

This identity is central. It shows that all candidate bindings are represented simultaneously using one shared $q$-qubit data register. The exponential search space is carried by the tuple register, not by duplicating the hypervector representation. \autoref{fig:tiny_example} depicts this idea in a small two-factor circuit.

\begin{examplebox}{Two-factor coherent binding}{coherentbinding}
Consider $F=2$, $N=2$, and $D=8$. The tuple register spans the four branches
\[
\ket{00},\ \ket{01},\ \ket{10},\ \ket{11}.
\]
After applying \(\mathrm{SELECT}_1\) and \(\mathrm{SELECT}_2\), the joint state becomes
\[
\frac{1}{2}
\bigl(
\ket{00}\ket{b_{00}}
+
\ket{01}\ket{b_{01}}
+
\ket{10}\ket{b_{10}}
+
\ket{11}\ket{b_{11}}
\bigr).
\]
For a fixed hyperdimension label $\ket{u}$, the effective sign may differ across tuple branches because the sign is attached to the joint basis state $\ket{\mathbf n}\ket{u}$, not to $\ket{u}$ alone.
\end{examplebox}

\subsection{Similarity Extraction by Walsh--Hadamard Overlap}

Once a candidate bound hypervector has been prepared on the data register, we evaluate its similarity with the target hypervector \(y\). Define the target phase oracle
\begin{equation}
U_y\ket{u}=y[u]\ket{u}.
\end{equation}
Acting on \(\ket{b_{\mathbf n}}\), this yields
\begin{equation}
U_y\ket{b_{\mathbf n}}
=
\frac{1}{\sqrt D}
\sum_{u=0}^{D-1}
y[u]\,b_{\mathbf n}[u]\ket{u}.
\end{equation}
Applying the Walsh--Hadamard transform \(H^{\otimes q}\) maps the average of these signed amplitudes to the basis state \(\ket{0^q}\). The amplitude on \(\ket{0^q}\) is therefore
\begin{equation}
\delta_{\mathbf n}
=
\frac{1}{D}
\sum_{u=0}^{D-1}
y[u]\,b_{\mathbf n}[u].
\end{equation}
This is exactly the HDC similarity score. Hence the decomposition score is encoded as an amplitude that can be interrogated by quantum search subroutines. \autoref{fig:overlap_extraction} shows this similarity-extraction circuit.

\begin{examplebox}{Exact similarity extraction}{overlapextraction}
Let \(D=4\), let
\[
y=[+1,-1,+1,-1],
\]
and first consider
\[
b_{\mathbf n}=[+1,-1,+1,-1].
\]
Then \(\delta_{\mathbf n}=1\), and after applying \(U_y\) followed by \(H^{\otimes 2}\), all amplitude concentrates on \(\ket{00}\). If instead
\[
b_{\mathbf n}=[+1,+1,-1,-1],
\]
then \(\delta_{\mathbf n}=0\), and the amplitude on \(\ket{00}\) vanishes. The similarity-extraction circuit therefore separates correct and incorrect candidates through amplitude concentration.
\end{examplebox}

\subsection{Maximum Finding via Modified Dürr--Høyer Quantum Search}

The similarity-extraction circuit yields a score oracle. We now embed it in a modified Dürr--Høyer maximum-finding procedure. The modification lies not in the outer logic of Dürr--Høyer itself, but in the way the comparison score is produced and marked: our score oracle is implemented through a logarithmic-qubit hypervector circuit rather than through explicit hypervector-state storage.

\medskip\noindent\textbf{Threshold oracle.}
Let \(\tau\in[-1,1]\) be the current threshold. Define the Boolean predicate
\begin{equation}
f_\tau(\mathbf n)
=
\begin{cases}
1, & \delta_{\mathbf n}>\tau,\\
0, & \delta_{\mathbf n}\le \tau.
\end{cases}
\end{equation}
The corresponding phase-marking oracle is
\begin{equation}
O_\tau\ket{\mathbf n}
=
(-1)^{f_\tau(\mathbf n)}\ket{\mathbf n}.
\end{equation}
At the circuit level, this oracle must be implemented reversibly. The score is first computed into workspace, then compared against \(\tau\), then converted into a phase flip, and finally uncomputed:
\begin{equation}
\ket{\mathbf n}\ket{0}
\longrightarrow
\ket{\mathbf n}\ket{f_\tau(\mathbf n)}
\longrightarrow
(-1)^{f_\tau(\mathbf n)}\ket{\mathbf n}\ket{f_\tau(\mathbf n)}
\longrightarrow
(-1)^{f_\tau(\mathbf n)}\ket{\mathbf n}\ket{0}.
\end{equation}
The final uncomputation is necessary to avoid residual entanglement with work registers. \autoref{fig:threshold_oracle} shows this compute--phase--uncompute structure.

\begin{examplebox}{Threshold marking}{thresholdmarking}
Suppose three candidate tuples have similarities
\[
\delta_{\mathbf n_1}=0.82,
\qquad
\delta_{\mathbf n_2}=0.41,
\qquad
\delta_{\mathbf n_3}=0.76.
\]
If the threshold is \(\tau=0.75\), then
\[
f_\tau(\mathbf n_1)=1,
\qquad
f_\tau(\mathbf n_2)=0,
\qquad
f_\tau(\mathbf n_3)=1.
\]
The threshold oracle therefore flips phase for \(\mathbf n_1\) and \(\mathbf n_3\), while leaving \(\mathbf n_2\) unchanged.
\end{examplebox}

\medskip\noindent\textbf{Maximum finding procedure.}
We now embed the threshold oracle into the Dürr--Høyer framework. The procedure maintains an incumbent tuple \(\mathbf n_{\mathrm{best}}\), uses its score as the current threshold, and searches for a tuple whose score exceeds that threshold. If one is found, the incumbent is updated; if not, the current incumbent is retained. Each improvement step is implemented by amplitude amplification using the oracle \(O_\tau\) and the diffusion operator
\begin{equation}
D_A=2\ket{A}\!\bra{A}-I.
\end{equation}
\autoref{fig:grover_iteration} shows one such iteration.
\
The resulting algorithm is best understood as modified Dürr--Høyer maximum finding rather than as a separate Grover search procedure. Grover-style amplification appears as an internal subroutine, but the outer logic is a maximum-finding routine over tuple scores.

The full pipeline is summarized in \autoref{alg:main}. The figures show the circuit structure of the core subroutines, while the pseudocode summarizes how these subroutines are assembled into the complete decomposition procedure.

\begin{algorithm}[t]
\caption{Qubit-Efficient HDC Decomposition}
\label{alg:main}
\begin{algorithmic}[1]
\STATE Prepare the tuple register in \(\ket{A}=\frac{1}{\sqrt{N^F}}\sum_{\mathbf n}\ket{\mathbf n}\)
\STATE Prepare the data register in \(\ket{+}^{\otimes q}\)
\STATE Construct the tuple-conditioned binding circuit \(\mathrm{BIND}\)
\STATE Construct the target overlap circuit using \(U_y\) and \(H^{\otimes q}\)
\STATE Initialize an incumbent tuple \(\mathbf n_{\mathrm{best}}\)
\REPEAT
\STATE Set \(\tau \leftarrow \delta_{\mathbf n_{\mathrm{best}}}\)
\STATE Build the threshold oracle \(O_\tau\)
\STATE Apply modified Dürr--Høyer search to find an improving tuple
\STATE Update \(\mathbf n_{\mathrm{best}}\) if an improving tuple is found
\UNTIL{termination}
\STATE Measure the tuple register and return \(\mathbf n_{\mathrm{best}}\)
\end{algorithmic}
\end{algorithm}

\section{Experimental Results}

We evaluate the proposed logarithmic hypervector and logarithmic binding encoding-based quantum framework from three complementary perspectives: \textbf{(i) \emph{whether the circuit computes the intended HDC similarity score correctly}}, \textbf{(ii) \emph{whether the resulting decomposition pipeline reliably recovers the correct tuple across a broad range of problem settings}}, and \textbf{(iii) \emph{whether the proposed representation provides favorable qubit-scaling behavior relative to the prior explicit hypervector encoding-based quantum baseline}}.

\subsection{Experimental Setup}

All experiments use synthetic dense bipolar codebooks with entries in \(\{-1,+1\}^{D}\). For each trial, we randomly generate \(F\) codebooks, each containing \(N\) candidate hypervectors, sample one codeword from each codebook, and form the target hypervector by bipolar binding. When noise is considered, we corrupt the target by randomly flipping a prescribed fraction of coordinates. This produces controlled decomposition instances with known ground-truth tuples.
\
For executable quantum experiments, we implement the proposed HDC decomposition circuit in Qiskit~\cite{javadi2024quantum} and simulate it with Qiskit Aer. We use both statevector simulation, to verify exact amplitude behavior, and shot-based simulation, to evaluate finite-sampling effects under realistic measurement budgets. The primary shot counts used in the experiments are \(256\), \(1024\), and \(2048\), depending on each evaluation. For decomposition experiments, we evaluate small-to-moderate settings that remain executable under repeated circuit simulation, including factors \(F\in\{2,3,4,5\}\), codebook sizes \(N\in\{2,3,4,5\}\), and hypervector dimensions \(D\in\{8,16,32,64\}\), subject to executable search-space limits.

\subsection{Correctness of the Similarity-Extraction Circuit} 

\begin{figure}[t]
    \centering
    \includegraphics[width=\linewidth]{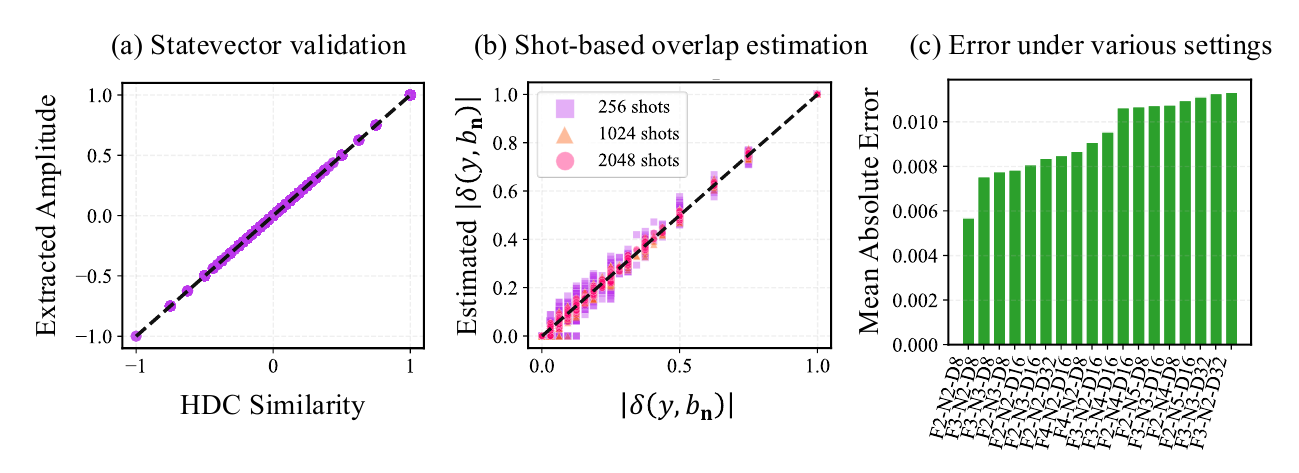}
    \vspace{-6mm}
    \caption{
    Correctness of the proposed logarithmic-qubit similarity-extraction circuit.
    \textbf{(a)} Exact HDC similarity versus the amplitude extracted from statevector simulation, showing near-perfect agreement with the identity line.
    \textbf{(b)} Exact similarity magnitude versus shot-based estimates for multiple shot budgets.
    \textbf{(c)} Dense setting-wise summary of estimation error across representative \((F,N,D)\) configurations.
    Across all panels, the results confirm that the proposed circuit computes the intended similarity signal and that measurement error decreases as the shot budget increases.
    }
    \label{fig:exp_similarity_correctness}
\end{figure}

\autoref{fig:exp_similarity_correctness} evaluates whether the proposed logarithmic hypervector encoding-based similarity-extraction circuit correctly computes the intended HDC similarity score. In \autoref{fig:exp_similarity_correctness}(a), the amplitude extracted from exact statevector simulation lies almost perfectly on the identity line when plotted against the analytical HDC similarity. This confirms that the circuit implements the desired similarity computation exactly at the wavefunction level. Equivalently, the amplitude associated with \(\ket{0^q}\) matches the theoretical score \(\delta(y,b_{\mathbf n})\), consistent with the construction described in the methodology.
\
\autoref{fig:exp_similarity_correctness}(b) shows that this agreement is preserved under finite-shot simulation. Although measurement noise introduces dispersion around the ideal line, the shot-based estimates become increasingly concentrated near the exact similarity magnitude as the number of shots increases. 
\
\autoref{fig:exp_similarity_correctness}(c) further summarizes the estimation error across a broad range of experimental settings, showing that the circuit remains accurate and stable over diverse decomposition configurations. Overall, these results provide strong empirical evidence that the proposed circuit faithfully and reliably computes the intended HDC similarity score.

\subsection{Decomposition Accuracy Across Problem Settings} 

\begin{figure}[t]
    \centering
    \includegraphics[width=\linewidth]{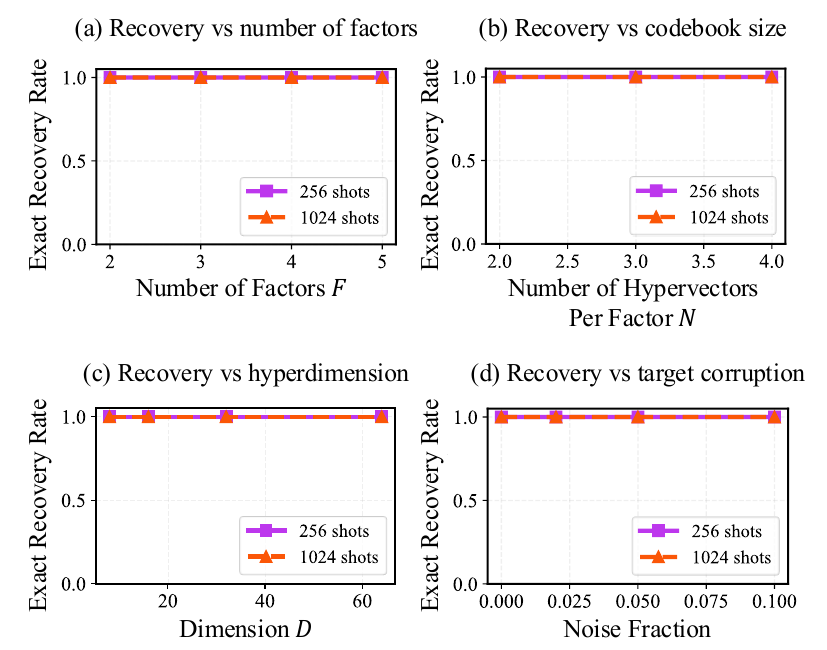}
    \vspace{-6mm}
    \caption{
    Decomposition accuracy of the proposed method across problem settings.
    \textbf{(a)} Exact recovery rate versus the number of factors \(F\).
    \textbf{(b)} Exact recovery rate versus the number of candidate hypervectors per factor \(N\).
    \textbf{(c)} Exact recovery rate versus the hypervector dimension \(D\).
    \textbf{(d)} Exact recovery rate under increasing target corruption.
    Shaded regions indicate one standard deviation across repeated trials.
    }
    \label{fig:exp_decomposition_accuracy}
\end{figure}

We next evaluate the end-to-end decomposition accuracy of the proposed method over a broad range of executable settings, as summarized in \autoref{fig:exp_decomposition_accuracy}. In particular, we vary the number of factors \(F\), the number of candidate hypervectors per factor \(N\), and the hypervector dimension \(D\), with the corresponding results shown in \autoref{fig:exp_decomposition_accuracy}(a), \autoref{fig:exp_decomposition_accuracy}(b), and \autoref{fig:exp_decomposition_accuracy}(c), respectively. Across all tested settings, the proposed method achieves \(100\%\) exact recovery, demonstrating that the logarithmic-qubit construction preserves the correctness of HDC decomposition over diverse problem configurations.
\
\autoref{fig:exp_decomposition_accuracy}(d) further examines robustness under increasing target corruption. The proposed method maintains \(100\%\) exact recovery under moderate noise levels, indicating that the decomposition procedure remains stable even when the target hypervector is partially corrupted.
\
Taken together, these results show that the proposed framework is not only correct in ideal settings, but also accurate and robust across a broad range of decomposition regimes and moderate levels of target corruption.

\begin{figure}[t]
    \centering
    \includegraphics[width=\linewidth]{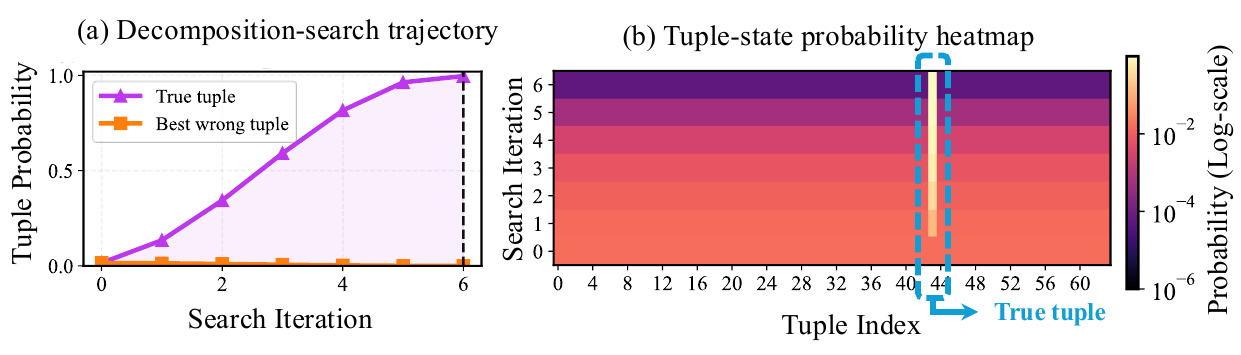}
    \vspace{-6mm}
    \caption{
    Decomposition-specific search dynamics under the modified Dürr--Høyer procedure.
    \textbf{(a)} Probability trajectory of the true decomposition tuple and the strongest incorrect tuple for a representative instance.
    \textbf{(b)} Heatmap of the full tuple-state probability distribution across search iterations.
    The plots show that probability mass is amplified toward the correct decomposition tuple and that the separation between the correct and strongest incorrect candidates increases over the search process.
    }
    \label{fig:exp_decomposition_search_evolution}
\end{figure}

We further examine the search dynamics of the proposed method on actual decomposition instances in \autoref{fig:exp_decomposition_search_evolution}.
\
\autoref{fig:exp_decomposition_search_evolution}(a) shows that the probability of the true tuple increases steadily across search iterations, while the strongest incorrect tuple remains consistently lower. This is precisely the behavior intended by the modified Dürr--Høyer procedure: amplitude is progressively concentrated on tuples whose scores exceed the current threshold, and when the correct tuple is the unique highest-scoring candidate, its probability should become dominant after an appropriate number of iterations.
\
\autoref{fig:exp_decomposition_search_evolution}(b) provides a more complete view of this evolution through a heatmap over the tuple-state distribution. Most candidate tuples remain at very low probability throughout the search, whereas the true tuple is selectively amplified. The highlighted reference box marks the correct tuple and shows that the dominant probability mass aligns with the intended decomposition candidate, rather than diffusing arbitrarily across the search space. These results provide direct evidence that the proposed search procedure behaves as expected on the decomposition problem itself, not merely on an abstract search benchmark.

\subsection{Qubit Efficiency}

\begin{figure}[t]
    \centering
    \includegraphics[width=\linewidth]{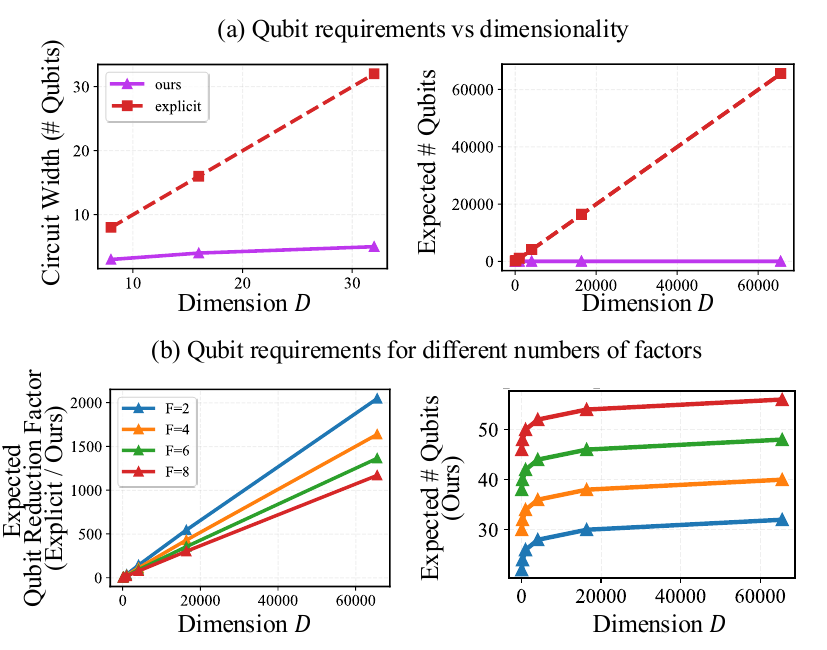}
    \vspace{-6mm}
    \caption{
Qubit requirement comparison between the proposed logarithmic hypervector encoding and an explicit hypervector encoding-based quantum baseline across different hyperdimensions \(D\). For small, executable settings, we report the actual transpiled circuit width. For larger values of \(D\), where full circuit simulation is no longer practical, we report the theoretically derived \emph{expected} qubit count.
\textbf{(a)} Actual transpiled circuit width on executable instances (left) and theory-driven qubit scaling based on the expected number of required qubits (right).
\textbf{(b)} Fine-grained qubit scalability analysis across different numbers of factors \(F\), including the qubit reduction factor achieved by the proposed method (left) and the proposed qubit requirement across hypervector dimensions for different values of \(F\) (right).
These results highlight the central qubit-efficiency advantage of our framework.
    }
    \label{fig:exp_qubit_requirements}
\end{figure}

\autoref{fig:exp_qubit_requirements} directly evaluates the central architectural claim of the paper: the proposed framework is substantially more qubit-efficient than explicit-dimension quantum encodings. As shown in the left panel of \autoref{fig:exp_qubit_requirements}(a), even the executable circuit instances already require fewer qubits under the proposed design than under the explicit baseline. The scalability gap becomes far more pronounced in the larger, theory-driven analysis. In the right panel of \autoref{fig:exp_qubit_requirements}(a), the required number of qubits under the proposed method grows only logarithmically with the hypervector dimension \(D\), whereas the explicit baseline grows linearly in \(D\).
\
This distinction is especially important in the high-dimensional regime where HDC is most useful. The left panel of \autoref{fig:exp_qubit_requirements}(b) shows that the qubit reduction factor increases rapidly as \(D\) grows, reaching approximately \(2{,}000\times\) in the evaluated range. The right panel of \autoref{fig:exp_qubit_requirements}(b) further illustrates that the proposed method remains compact even as the problem dimension increases, whereas the explicit encoding quickly becomes impractical. Overall, these results provide strong evidence that the proposed logarithmic-qubit design resolves a fundamental scalability limitation of prior explicit hypervector encodings.

\section{Conclusion}
We presented a qubit-efficient quantum framework for hyperdimensional decomposition based on logarithmic hypervector encoding, logarithmic binding encoding, and an explicit hypervector lookup operator. The proposed method preserves the \(\mathcal{O}(\sqrt{N^F})\) quantum search complexity of prior work while reducing hypervector representation cost from \(\mathcal{O}(D)\) qubits to \(\mathcal{O}(\log D)\). Experiments verified correct similarity computation, accurate decomposition in executable regimes, and substantial qubit savings, reaching approximately \(2{,}000\times\) reduction in the evaluated high-dimensional setting. These results show that qubit-efficient circuit design is essential for making quantum HDC decomposition meaningfully closer to practical implementation.

\begin{acks}
This work was supported in part by the DARPA Young Faculty Award, the National Science Foundation (NSF) under Grants \#2127780, \#2319198, \#2321840, \#2312517, and \#2235472, the Semiconductor Research Corporation (SRC), the Office of Naval Research through the Young Investigator Program Award, and Grants \#N00014-21-1-2225 and N00014-24-1-2547, Army Research Office Grant \#W911NF2410360. Additionally, support was provided by the Air Force Office of Scientific Research under Award \#FA9550-22-1-0253.
\end{acks}

\bibliographystyle{ACM-Reference-Format}
\bibliography{mybibliography}










\end{document}